\documentclass[journal]{IEEEtran}
\usepackage{verbatim}
\usepackage{amsfonts}

\usepackage{graphicx}
\usepackage{amsmath,amssymb} 
\usepackage{color}
\usepackage{epstopdf}

\usepackage{float}
\usepackage[colorlinks,linkcolor=black]{hyperref}
\usepackage{multirow}
\usepackage{makecell}
\usepackage{color}
\usepackage{graphicx}
\ifCLASSINFOpdf
\else
\fi

\begin{document}
\title{Class balanced underwater object detection dataset generated by class-wise style augmentation}
\author{
Long~Chen~
Junyu Dong,~
Huiyu Zhou
\thanks{L. Chen and H. Zhou are with School of Informatics, University of Leicester, United Kingdom, e-mail: (lc408, hz143@leicester.ac.uk).}
\thanks{J. Dong with Department of information science and engineering, Ocean University of China, China, e-mail: (dongjunyu@ouc.edu.cn).}
}

\markboth{}
{Shell \MakeLowercase{\textit{et al.}}: Bare Demo of IEEEtran.cls for IEEE Journals}

\maketitle
\begin{abstract}
Underwater object detection technique is of great significance for various applications in underwater the scenes. However, class imbalance issue is still a unsolved bottleneck for current underwater object detection algorithms. It leads to large precision discrepancies among different classes that the dominant classes with more training data achieve higher detection precisions while the minority classes with fewer training data achieves much lower detection precisions. In this paper, we propose a novel class-wise style augmentation (CWSA) algorithm to generate a class-balanced underwater dataset Balance18 from the public contest underwater dataset URPC2018. CWSA is a new kind of data augmentation technique which augments the training data for the minority classes by generating various colors, textures and contrasts for the minority classes. Compare with previous data augmentation algorithms such flipping, cropping and rotations, CWSA is able to generate a class balanced underwater dataset with diverse color distortions and haze-effects. 

\end{abstract}
%
\IEEEpeerreviewmaketitle
\section{Introduction}
Large-scale datasets with high quality annotations are of vital important for training deep neural networks (DNNs), especially in the underwater object detection task where large amounts data access is limited and the annotation of the data is often expensive~\cite{b1, b2, b3}. A common solution is to augment smaller datasets by creating new training samples from existing ones via label-preserving transformation. 

Typical data augmentation techniques~\cite{b4, b5, b6} include image cropping, flipping and warping and other deformations to create the augmented images. More complex techniques can include noise addition, geometric transformations, or image
compression. Previous works have shown data augmentation is able to boost the performance of many computer vision tasks such as image classification~\cite{b7, b8, b9, b10, b11}, object detection~\cite{b12} and image segmentation~\cite{b13, b14}. However, there two limitations of when applying these data augmentation methods in underwater scenes. First, underwater images suffer from a diverse of visual degradations such as various color distortion and haze effects~\cite{b15, b16, b17, b18}, and typical data augmentation methods can generate these diverse images which limit the generalisation of the deep model on the real-world applications. Second, previous data augmentation methods are not designed for generating a class-imbalanced dataset, hence the deep detectors trained on them achieve very low precisions for the minority classes whose training samples are insufficient~\cite{b19}. In this paper, firstly, we apply the unsupervised CycleGAN~\cite{b20} framework to conduct style transfer on the underwater dataset to generate diverse underwater images. Second, we improve the style transfer method as a class-wise style augmentation method to generate a class-balanced underwater dataset. The objective of this work is to use a style transfer network as a generative model to create more samples for minor classes. Since style transfer preserves the overall semantic content of the original image, the high-level discriminative features of an object are maintained. To the best of our knowledge, we are the first to use style transfer for data augmentation to address the class imbalance problem.

\begin{figure*}[h]
\centering
\includegraphics[height=6cm, width=18cm]{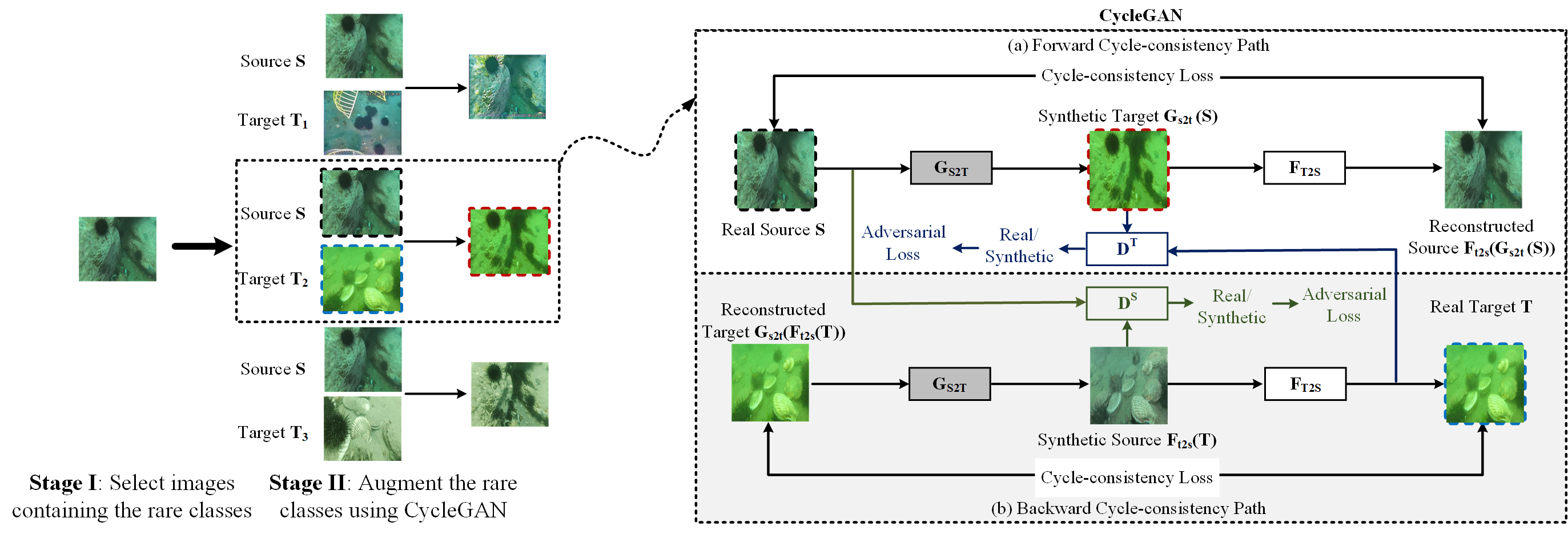}
\caption{The proposed class-wise style augmentation method. It consists of two stage: 1) Select images with more instances of the minority categories; (2) Conduct data augmentation on the selected images to augment the minority categories.}
\label{fig:CWDA}
\end{figure*}
Our unique contributions can be summarised as follows:
\begin{itemize}
\item We propose a class-wise style augmentation algorithm to augment underwater images with various styles, it fully takes into account underwater image characteristics such as color distortion, haze-effects. To our knowledge, this is the first practice for data augmentation, aiming to augment images with various color style and blurry style.
\item The style augmentation is class-wise, which aims to generate a balanced underwater dataset. The augmented dataset can be used to alleviate the class-imbalance issue in underwater recognition tasks.
\end{itemize}

The rest of the paper is organised as follows: Section \ref{sec:relatework} summarises the related works. Section \ref{sec:proposedmethod} describes the proposed class-wise style augmentation algorithm. Section \ref{sec:exmperiments} describes the experimental set-up and reports the experimental results.

\section{Related Work}
\label{sec:relatework}
\subsection{Data Augmentation}
Data augmentation~\cite{b21} has been a standard and effective technique for improving the generalization of deep networks since the success of AlexNet in the ImageNet classification competition~\cite{b7}. It refers to the process of adding more variation to the training data in order to improve the generalisation capabilities of the trained model. It is particularly useful in scenarios such as underwater environments where there is a scarcity of training samples. Data augmentation imparts prior knowledge to a model by explicitly teaching invariance to possible transforms that preserve semantic content. This is done by applying the transformation to the original training data, producing new samples whose labels are known. For example, horizontal flipping is a popular data augmentation technique, as it clearly does not change the corresponding class label. The most prevalent forms of image-based data augmentation include geometric distortions such as random cropping, zooming, rotation, flipping, linear intensity scaling and elastic deformation. Whilst these are successful at teaching rotation and scale invariance to a model, but fail to incorporate color, texture and complex illumination variations. On the other hand, style augmentation is able to alter the distribution of low-level visual features such as color, illumination and contrast whilst preserving semantic content. Hence, in this paper, we propose a style augmentation method to augment the training data of minor classes (i.e., classes with less training instances). It generates different color, texture and contrast for the minor classes whilst preserving geometry.

\subsection{Style Transfer}
With the success of deep learning in numerous computer vision tasks, researchers have moved their focus from traditional data augmentation methods to data-driven augmentation methods. Style transfer algorithms~\cite{b22, b23, b24} aim to modify the visual style of an image while preserving its semantic content, and have shown their effectiveness in many computer vision tasks. Tobin et al.~\cite{b25} synthesize images by randomizing the color, texture, illumination and other aspects of the virtual scene and use the synthetic images to train a deep detector, they discover the detector generalizes well from graphically rendered virtual environments to the real world. They also observe the diversity of the synthetic images is more important than the realistic of them for the generalisation of the deep models. Inspired by this observation, Jackson et al.~\cite{b26} propose style augmentation to apply style transfer to augment arbitrary training images, randomizing their color, texture and contrast whilst preserving geometry. They use the randomized action of the style transfer pipeline to augment image datasets to greatly improve downstream model performance across a range of tasks. All these works demonstrate that neural style transfer algorithms can apply the artistic style (color, contrast, and texture) of one image to another image without changing the latter's high-level semantic content, which makes them feasible to employ neural style transfer as a data augmentation method to add more variation to the training dataset.

\section{Proposed methods}
\label{sec:proposedmethod}
\begin{figure*}[htb]
\centering
\includegraphics[height=4.5cm, width=14cm]{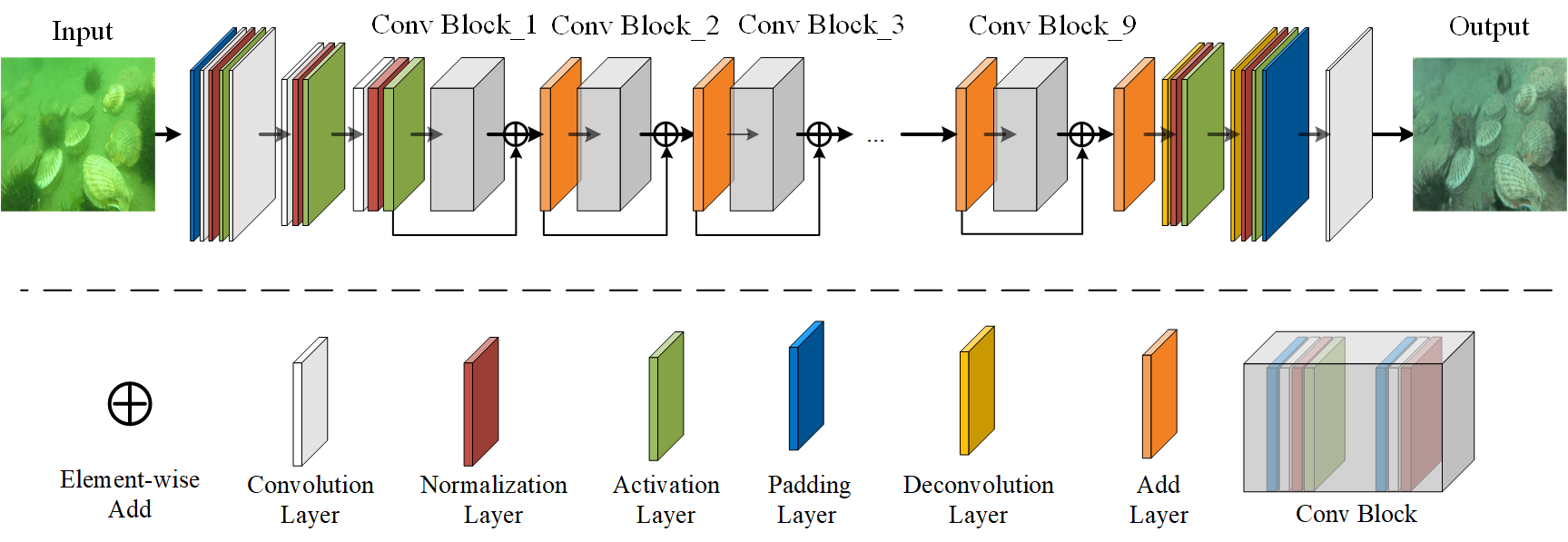}
\caption{The structure of generator in CycleGAN.}
\label{fig:generator}
\end{figure*}

In this section, we first describe the real-world datasets we used in this work. Then, we present the overview of our proposed class-wise style augmentation framework. Finally we describe CycleGAN in detail. 
\subsection{Underwater datasets}

In our work, we use two public underwater robot picking contest datasets URPC2017 and URPC2018\footnote{The dataset can be downloaded from http://www.cnurpc.org/index.html}, which are provided by National Natural Science Foundation of China and Dalian Municipal People’s Government. The contest holds annually from 2017, consisting of online and offline object detection contests.  Two datasets we use here come from the online object detection contest. Two datasets both suffer from severe class imbalance problem, in addition, the annotations of URPC2017 contains considerable noisy labels that make it is not suitable for evaluating  different object detection algorithms fairly. Hence, we only conduct data augmentation on the real-world underwater dataset URPC2018, which contain much less noisy labels. We manually divides all the images from URPC2017 and URPC2018 into four categories: green, blue, deepblue and white. Then, for each category, we select the same numbers of images to form a balanced underwater dataset. This is because the style transfer framework CycleGAN is also influenced by the class imbalance problem, it is unable to generate realistic enough synthetic images for the minority classes. We combine two datasets, because only one dataset cannot form a large enough balanced dataset for training the CycleGAN.

\subsection{Class-wise Style Augmentation Framework}

We present our proposed class-wise style augmentation in this section. We explore the state-of-the-art neural style transfer algorithms and apply them as a data augmentation method for minority classes in the underwater datasets. As shown in Figure~\ref{fig:CWDA}, we first select the images containing more instances of the minority categories. For example, in the URPC2018 dataset, the scallop and seacucumber categories have much less object instances than the seaurchin and starfish categories. Hence, we select the images containing the scallops or sea cucumbers most and  remove other images. Then, we conduct data augmentation only on these selected images to augment the minority classes. Our proposed class-wise data augmentation methods only augment the minority categories, thus it is able to generate a balance underwater dataset for all categories.  In our works, we apply the CycleGAN, a type of unsupervised generative adversarial network to conduct style transfer. It allows the mutual transformation between different styles with only unpaired images.
\begin{figure}[htb]
\centering
\includegraphics[height=4.5cm, width=8cm]{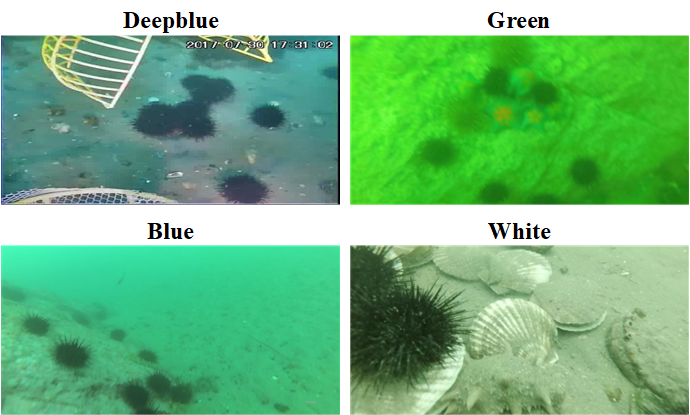}
\caption{The underwater images in URPC2017 and URPC2018 can be divided into four color categories.}
\label{fig:imexemplars}
\end{figure}
\begin{figure}[htb]
\centering
\includegraphics[height=6.0cm, width=9cm]{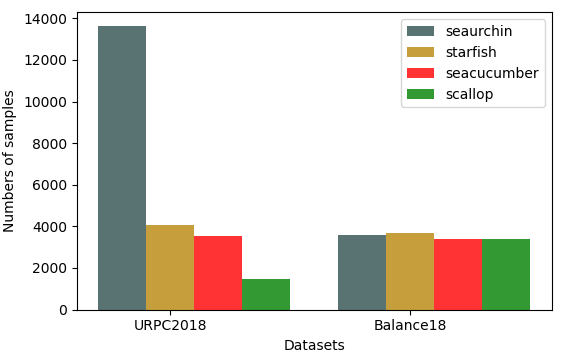}
\caption{The data distribution for all categories on URPC2018 and Balance18.}
\label{fig:class}
\end{figure}
\begin{figure*}[htb]
\centering
\includegraphics[height=8cm, width=16cm]{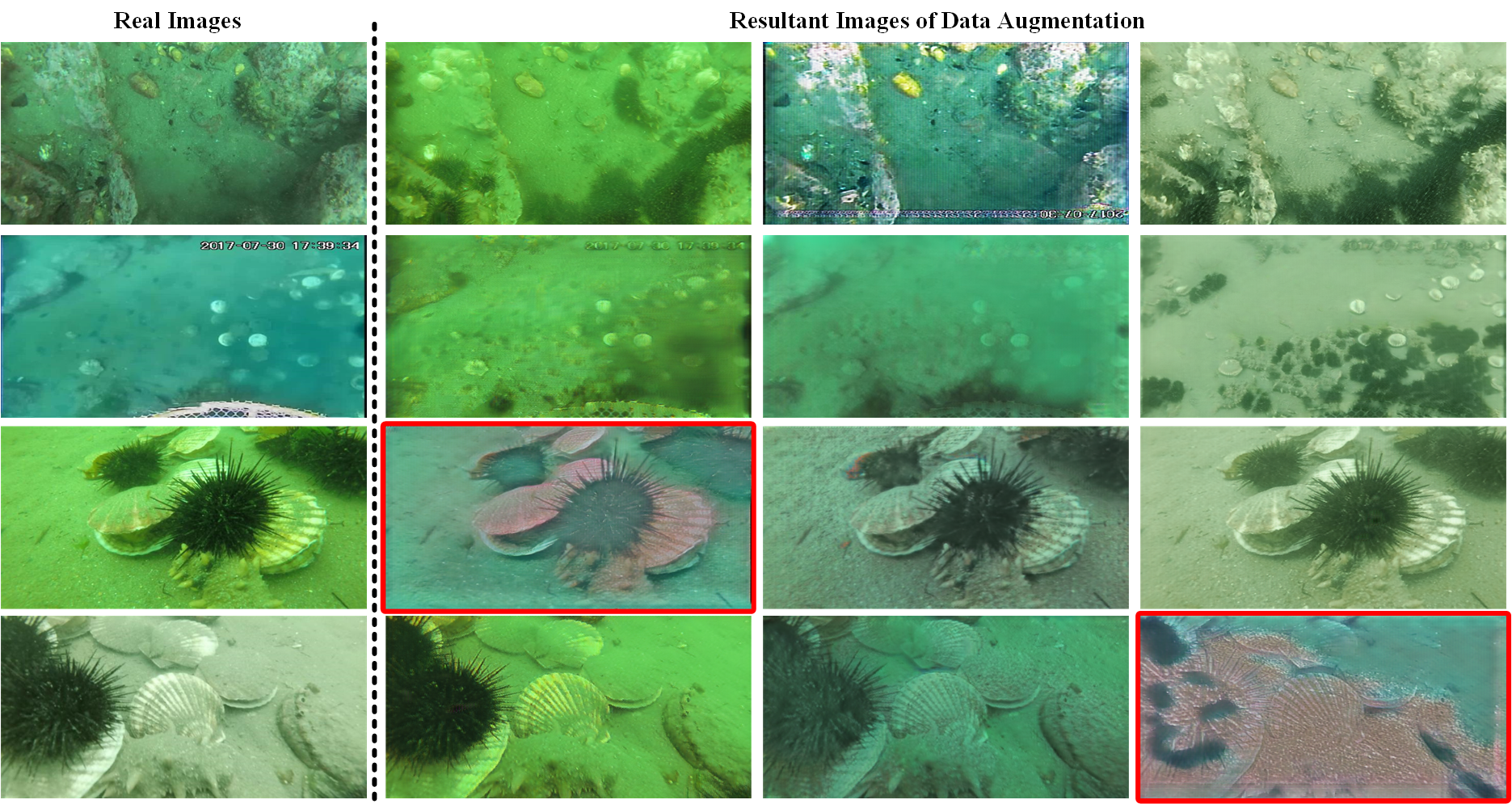}
\caption{Augmented images for the real-world underwater images. The first column are the original images, the other columns are the augmented images, the unsatisfactory resultant image (red boxes) have been checked and deleted from the Balance18 dataset.}
\label{fig:results}
\end{figure*}

\subsection{Style Transfer framework CycleGAN}

Cycle-Consistent GAN (CycleGAN) is a type of generative adversarial network for unpaired image-to-image translation. Unlike other GAN models for image translation, it does not require a dataset of paired images. This allows the development of a translation model on problems where training datasets may not exist, such as translating underwater images to clear images. Traditionally, training an image-to-image translation model requires a dataset comprised of paired examples. However, the requirement for a paired training dataset is a limitation. These datasets are challenging and expensive to prepare, e.g. photos of different scenes underwater different conditions. In many cases, the datasets simply do not exist. As shown in Figure 1, the architecture of CycleGAN is composed of four models, two discriminators and two generators. The discriminator is a deep convolutional neural network that performs image classification. It takes a source image as input and predict the likelihood of whether the target image is a real or fake image. The generator is an encoder-decoder model architecture. It takes a source image and generates a target image. As shown in Figure~\ref{fig:generator}, the generator does this by first down sampling the input image to a bottleneck layer, then up sampling the representation to the size of the output images. The discriminators are trained directly on real and generate images, whereas the generators are not. Instead, the generators are trained via their related discriminators. Specially, they are updated to minimize the loss predicted by the discriminator for generated images as real, called adversarial loss. As such, they are encouraged to generate images that better fit into the target domain. The generator is also updated based on how effectively it is at the regeneration of a source image, called cycle loss. Cycle loss are calculated as L1 distance between the input and output image for each sequence of translations. Adversarial loss is calculated as the L2 distance between the model output and the target values of 1.0 for real and 0.0 for fake.

\section{Experimental Setup}
\label{sec:exmperiments}
In this section, we first introduce the experimental datasets and implementation details. Then, we present the experimental results of the proposed class-wise data augmentation.

\subsection{Dataset and Implementation Details}

URPC2017 and URPC2018 datasets are two public competition dataset for evaluating underwater object detection algorithms and underwater image enhancement algorithms. The URPC2017 dataset has 3 object categories, including seacucumber, seaurchin and scallop. There are 18,982 training images and 983 testing images. The URPC2018 dataset has 4 object categories, including seacucumber, seaurchin, scallop and starfish. There are 2,897 images in the training set, since the testing set is not publicly available, we randomly split the training set of URPC2018 into a training set of 1,999 images and a testing set of 898 images. Both two datasets provide underwater images and box level annotations.

We generate a class balanced dataset Balance18 from the class imbalanced URPC2018 using the proposed class-wise data augmentation method. The comparison of data distribution on URPC2018 and Balance2018 are shown in Figure~\ref{fig:class}, from which we observe URPC2018 contains highly imbalanced classes, the seaurchin category has much more samples than other categories. After applying the CWSA on the imbalance URPC2018, we obtain a balanced dataset Balance18, all categories have similar numbers of samples.

\subsection{Results and discussion}

We present several images generated by CWSA in Figure.~\ref{fig:results}. For the input real underwater images, we choose the images with more minority classes and less majority classes. We manually divide the images into four categories according to the color distortion, and transform each category to other three categories to augment the minority classes. As shown in Figure.~\ref{fig:results}, the generated images are of diverse colors, textures and contrasts whilst preserving the geometry. However, there are some serious artefacts in several resultant images, hence we manually select and drop them. Applying CWSA to the selected images in URPC2018, we are able to generate a class-balanced underwater object detection dataset Balance18.

\section{Conclusion}
In this paper, we have proposed a class-wise style transfer algorithms to augment the minority classes. It generates a class-balanced dataset where images are of diverse colors, textures and haze-effects. The diversity of this dataset will benefit the DNNs by improving their generalisation on real-world underwater dataset. In addition, the balanced dataset can be used to alleviate the class-imbalance issue in real-world underwater recognition applications.

\end{document}